\documentclass[sigconf]{acmart}
\AtBeginDocument{%
  }

\copyrightyear{2026}
\acmYear{2026}
\setcopyright{cc}
\setcctype{by}
\acmConference[MM '26]{Proceedings of the 34th ACM International Conference on Multimedia}{November 10--14, 2026}{Rio de Janeiro, Brazil}
\acmBooktitle{Proceedings of the 34th ACM International Conference on Multimedia (MM '26), November 10--14, 2026, Rio de Janeiro, Brazil}
\acmDOI{10.1145/3767308.3835425}
\acmISBN{979-8-4007-2213-4/2026/11}

\usepackage{multirow}
\usepackage{enumitem}
\usepackage[table]{xcolor}
\usepackage{balance}
\definecolor{myblue}{RGB}{127, 200, 239}
\definecolor{mygreen}{RGB}{161, 215, 124}
\definecolor{myred}{RGB}{231, 51, 35}
\definecolor{idpdtgreen}{RGB}{72, 129, 35}
\definecolor{dpdtred}{RGB}{172, 79, 34}
\definecolor{R1}{RGB}{255, 0, 0}     % red
\definecolor{R2}{RGB}{0, 255, 0}     % green
\definecolor{R3}{RGB}{0, 0, 255}     % blue
\definecolor{R4}{RGB}{255, 128, 0}   % orange
\definecolor{R5}{RGB}{255, 0, 255}   % magenta

\begin{document}

%%
%% The "title" command has an optional parameter,
%% allowing the author to define a "short title" to be used in page headers.
\title{Filling the Unseen: Scene Extrapolation via 3D Gaussian Splatting}

%%
%% The "author" command and its associated commands are used to define
%% the authors and their affiliations.
%% Of note is the shared affiliation of the first two authors, and the
%% "authornote" and "authornotemark" commands
%% used to denote shared contribution to the research.
\author{Yunlai Zhou}
% \authornote{Both authors contributed equally to this research.}
% \orcid{0009-0004-7727-8657}
% \author{G.K.M. Tobin}
% \correspondingauthor
% \authornotemark[1]
% \email{webmaster@marysville-ohio.com}
\affiliation{%
  \institution{Case Western Reserve University}
  \department{Department of Computer and Data Sciences}
  \city{Cleveland}
  \state{Ohio}
  \country{USA}
}
\email{yxz3057@case.edu}

\author{Yiren Lu}
\affiliation{%
  \institution{Case Western Reserve University}
  \department{Department of Computer and Data Sciences}
  \city{Cleveland}
  \state{Ohio}
  \country{USA}
}
\email{yxl3538@case.edu}

\author{Tuo Liang}
\affiliation{%
  \institution{Case Western Reserve University}
  \department{Department of Computer and Data Sciences}
  \city{Cleveland}
  \state{Ohio}
  \country{USA}
}
\email{txl859@case.edu}

\author{Disheng Liu}
\affiliation{%
  \institution{Case Western Reserve University}
  \department{Department of Computer and Data Sciences}
  \city{Cleveland}
  \state{Ohio}
  \country{USA}
}
\email{dxl952@case.edu}

\author{Vipin Chaudhary}
\affiliation{%
  \institution{Case Western Reserve University}
  \department{Department of Computer and Data Sciences}
  \city{Cleveland}
  \state{Ohio}
  \country{USA}
}
\email{vxc204@case.edu}

\author{Yu Yin}
\correspondingauthor
\affiliation{%
  \institution{Case Western Reserve University}
  \department{Department of Computer and Data Sciences}
  \city{Cleveland}
  \state{Ohio}
  \country{USA}
}
\email{yxy1421@case.edu}

%%
%% By default, the full list of authors will be used in the page
%% headers. Often, this list is too long, and will overlap
%% other information printed in the page headers. This command allows
%% the author to define a more concise list
%% of authors' names for this purpose.
\renewcommand{\shortauthors}{Yunlai Zhou et al.}

%%
%% The abstract is a short summary of the work to be presented in the
%% article.
\begin{abstract}
3D Gaussian Splatting achieves photorealistic reconstruction within training view distribution, yet it degrades on out-of-distribution novel views, exhibiting holes in unobserved regions and artifacts in observable areas.
Recent works formulat this task as extrapolation and interpolation and tried to address it with generative models, but remain limited in extrapolation scale and quality.
They repeat a generate–reconstruct–shift cycle to progressively build a scene, which introduces accumulated errors with every step conditioning on previous outcomes.
In this work, we propose a holistic framework for extrapolation and interpolation.
We devise an independent camera view detection mechanism to enable parallel conflict-free extrapolation, circumventing the reliance on aforementioned error-prone cycle.
Building upon this, we design a hierarchical pipeline that extrapolates independent and dependent camera views separately.
Additionally, previous methods overlook inconsistency between generated and original images, resulting in compromising well-reconstructed areas.
We propose a plug-and-play Quality-Aware Mask (QA-Mask) module, enabling selective utilization on generated data. By calibrating learning weights with pixel-wise rendering quality, it prevents generation-induced degradations on well-constructed areas.
Extensive experiments demonstrate the superior performance of our framework, with QA-Mask generalizing on multiple generative reconstruction models (\href{https://vulab-ai.github.io/filling-the-unseen/}{project page}).
\end{abstract}

%%
%% The code below is generated by the tool at http://dl.acm.org/ccs.cfm.
%% Please copy and paste the code instead of the example below.
%%


%%
%% Keywords. The author(s) should pick words that accurately describe
%% the work being presented. Separate the keywords with commas.
%% A "teaser" image appears between the author and affiliation
%% information and the body of the document, and typically spans the
%% page.
% \begin{teaserfigure}
%  \includegraphics[width=\textwidth]{sampleteaser}
%   \caption{Seattle Mariners at Spring Training, 2010.}
%   \Description{Enjoying the baseball game from the third-base
%   seats. Ichiro Suzuki preparing to bat.}
%   \label{fig:teaser}
% \end{teaserfigure}

% \received{20 February 2007}
% \received[revised]{12 March 2009}
% \received[accepted]{5 June 2009}

%%
%% This command processes the author and affiliation and title
%% information and builds the first part of the formatted document.
\maketitle

\begin{figure}[t]
    \centering
    \includegraphics[width=\linewidth]{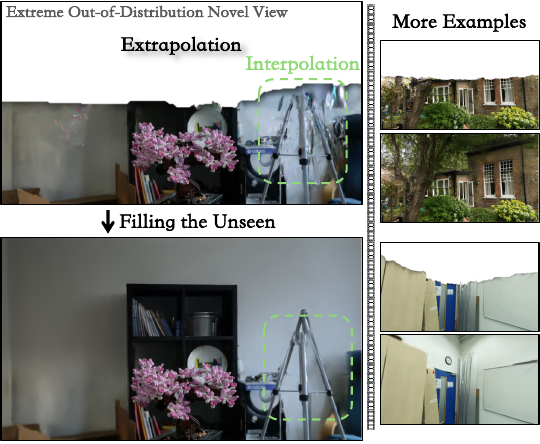}
    \caption{Our framework excels at large-scale, geometrically coherent 3D scene extrapolation and interpolation on extreme out-of-distribution novel views.}
    \label{fig:teaser}
\end{figure}

\vspace{-0.5em}
\section{Introduction}
\label{sec:intro}
The emergence of Neural Radiance Fields (NeRFs) \cite{mildenhall2021nerf} and 3D Gaussian Splatting (3DGS) \cite{kerbl20233d} has significantly advanced 3D scene reconstruction with their unprecedented photorealism and high fidelity. However, this ability is confined to the training view distribution. When rendering from out-of-distribution novel views, they suffer from severe degradations, including holes due to un-observation in training images, as well as artifacts and incorrect geometry in observed areas, as shown in the upper part of Figure~\ref{fig:teaser}. These issues undermine the viewing experience in image quality and freedom of exploration, limiting their practical applicability.

Several studies \cite{wu2025difix3d+, liu20243dgs, wei2025gsfix3d, yin2025gsfixer, lu2024view} employed diffusion models to repair artifacts on novel views, but they cannot handle holes due to un-observation in training views. Recently, a number of methods \cite{wu2025genfusion, zhong2025taming, hwang2024vegs, shih2024extranerf} have been proposed, formulating and trying to tackle the aforementioned two issues as scene extrapolation and interpolation. However, they are generally limited to small-scale extrapolation, such as filling small holes within observed areas, and are unable to expand a scene at a larger scale. Moreover, their extrapolation results are often prone to blurring and inaccurate geometry. We attribute these limitations to their reliance on the generate–reconstruct–shift cycle which is first introduced in text-to-scene generation \cite{zhang2024text2nerf, hollein2023text2room, ouyang2023text2immersion, chung2023luciddreamer, yu2025wonderworld}, where “shift” means moving the camera view to reveal new areas that need to be generated. This cycle will be iteratively repeated in a spatially-sequential manner to progressively build up a scene. However, iterative generation and reconstruction conditioned on the outcomes of previous steps tend to accumulate errors, as generation-based reconstruction is notoriously under-constrained, especially when extrapolating a scene beyond its boundaries.

Furthermore, existing approaches \cite{wu2025genfusion, zhong2025taming, hwang2024vegs, shih2024extranerf, wu2025difix3d+, liu20243dgs} overlook the inconsistency between generated content and the real scene. Indiscriminately applying generated content to update the scene inevitably leads to suboptimal reconstruction quality and fidelity, e.g., corrupting well-reconstructed areas in the original scene. A principled strategy is to selectively leverage generated content based on the current rendering quality, ensuring that generation only complements rather than undermines the reconstruction process.

In this paper, we present \textbf{\textit{Filling the Unseen}}, a holistic framework for large-scale, geometrically coherent scene extrapolation and interpolation via 3DGS, as shown in Figure~\ref{fig:teaser}. Our key novelty is that we switch the common paradigm from many-stage cycle repetition to hierarchical two-stage extrapolation. We first devise a mechanism to identify a set of independent camera views used for extrapolation in a given 3D scene by detecting collisions between mesh-based view frustums. These independent camera views support parallel and conflict-free generation and reconstruction. Built upon this mechanism, we redesign a hierarchical pipeline that circumvents repeating the error-prone generate–reconstruct–shift cycles. Specifically, our pipeline first performs large-scale scene extrapolation on all independent camera views in a single pass, followed by a second-stage residual completion for remaining unfilled regions, thus achieving seamless scene extrapolation. In addition, to improve the rendering quality of the original scene on out-of-distribution extrapolation views, we perform scene interpolation by re-exploiting the informative generated videos produced during the extrapolation, together with an artifact removal model \cite{wu2025difix3d+}.

Furthermore, by leveraging the directional modeling capability of spherical harmonics to model "which viewing directions are supervised" for each Gaussian primitive, we design a novel method to render masks that can measure rendering quality. Spatially re-weighting the reconstruction loss with this mask during scene updates enables selective utilization on generated content, as shown in Figure~\ref{fig:why_oa_mask}, which effectively prevents generation-induced degradations in well-reconstructed regions while also preserves desired extrapolation and interpolation. We show that this plug-and-play \textbf{Q}uality-\textbf{A}ware \textbf{Mask} (QA-Mask) module can be readily integrated into multiple generative reconstruction models \cite{wu2025difix3d+, wu2025genfusion} as enhancement. In summary, our key contributions include:
\begin{itemize}
    \item We propose \textit{Filling the Unseen}, a hierarchical 3DGS-based framework enabled by independent camera view detection mechanism for large-scale, geometrically coherent scene extrapolation and interpolation.
    \item We introduce QA-Mask, a plug-and-play, rendering quality-aware mask method to enable selective utilization on generated content during scene updates.
    \item Extensive experiments on multiple datasets demonstrate the superior performance of our proposed framework and the generalizability of QA-Mask.
\end{itemize}

\section{Related Works}
\subsection{Scene Extrapolation and Interpolation}
ExtraNeRF \cite{shih2024extranerf} first explored diffusion inpainters
to synthesize beyond observed regions, but was mainly limited to narrow regions around object boundaries. VEGS \cite{hwang2024vegs} extended this idea to urban driving scenes. For sparse inputs, Zhong et al. \cite{zhong2025taming} proposed a training-free scene-grounding strategy that tames pretrained diffusion models toward generating scene-consistent image sequences. GenFusion \cite{wu2025genfusion} instead trains a video diffusion model on paired low-quality renderings and ground-truth images, providing a unified solution to extrapolation and interpolation, but often suffering from blurriness and color inconsistency. GaMo \cite{huang2025gamo} formulates sparse-view reconstruction as multi-view outpainting by expanding the field of view to expand scene coverage.

Compared to extrapolation, scene interpolation has been more extensively studied. 3DGS-Enhancer \cite{liu20243dgs} fine-tunes diffusion models on held-out degraded renderings and their ground truth. DiFix3D+ \cite{wu2025difix3d+} improves this paradigm using single-step diffusion, diverse data hold-out strategies, and progressive 3D updates. GSFix3D \cite{wei2025gsfix3d} combines meshes and 3D Gaussians to handle diverse scenes and artifacts, while GSFixer \cite{yin2025gsfixer} incorporates VGGT \cite{wang2025vggt} into video diffusion models to condition generation on both geometry and appearance. Nevertheless, these approaches rely explicitly or implicitly on iterative generate-reconstruct-shift cycles and insufficiently address inconsistencies between generated content and real scenes, both of which require immediate and careful handling.

\subsection{Rendering Quality Measurement}
Rendering quality and uncertainty estimation benefit applications including next-best-view selection \cite{pan2022activenerf,xue2024neural}, floater removal \cite{goli2024bayes,warburg2023nerfbusters,li2024variational}, pruning and densification \cite{hanson2025pup}, and scene generation \cite{liu20243dgs}. For NeRFs, prior works model uncertainty through volumetric fields \cite{goli2024bayes} via spatial perturbations and Bayesian Laplace approximation, local 3D diffusion priors \cite{warburg2023nerfbusters} and score distillation sampling loss, or Neural Visibility Fields \cite{xue2024neural}. For Gaussian Splatting, Li et al. \cite{li2024variational} infer predictive uncertainty using a variational multi-scale framework, while PUP-3DGS \cite{hanson2025pup} uses second-order reconstruction-error approximations for uncertainty-aware pruning.

However, these methods are not designed for perceptual rendering quality assessment in context of scene generation. 3DGS-Enhancer \cite{liu20243dgs} assumes that well-reconstructed regions are represented by small-volume Gaussians, which may fail for large, low-frequency surfaces such as plain walls. In contrast, our QA-Mask better reflects perceptual rendering quality and is specifically designed for scene generation, as validated in our experiments.

\begin{figure}
    \centering
    \includegraphics[width=\linewidth]{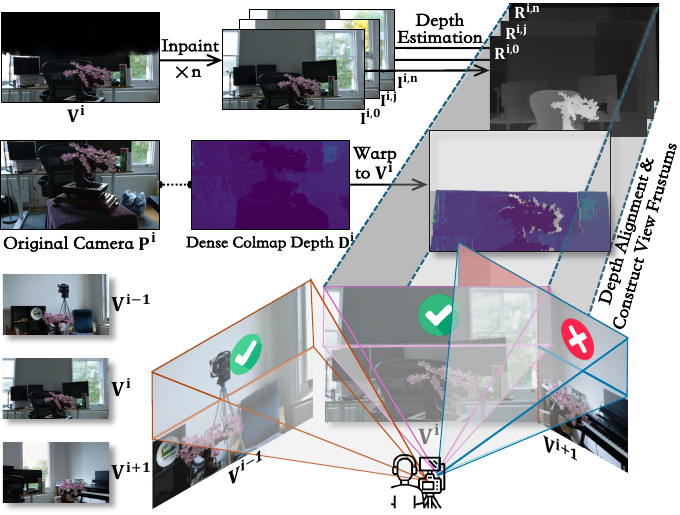}
    \caption{\textbf{Independent Camera View Detection.} We construct a mesh-based view frustum enclosing the geometry of the inpainted region for each candidate view $\mathbf{V}^{\text{i}}$. Collision detection is performed for frustums to determine whether their camera views are independent. In this example, $\mathbf{V}^{\text{i-1}}$ and $\mathbf{V}^{\text{i}}$ are independent, while $\mathbf{V}^{\text{i}}$ and $\mathbf{V}^{\text{i+1}}$ collide with each other and are therefore not independent.}
    \label{fig:independent_camera_view_detection}
\end{figure}

\section{Methods}
\begin{figure*}[t]
    \centering
    \includegraphics[width=\linewidth]{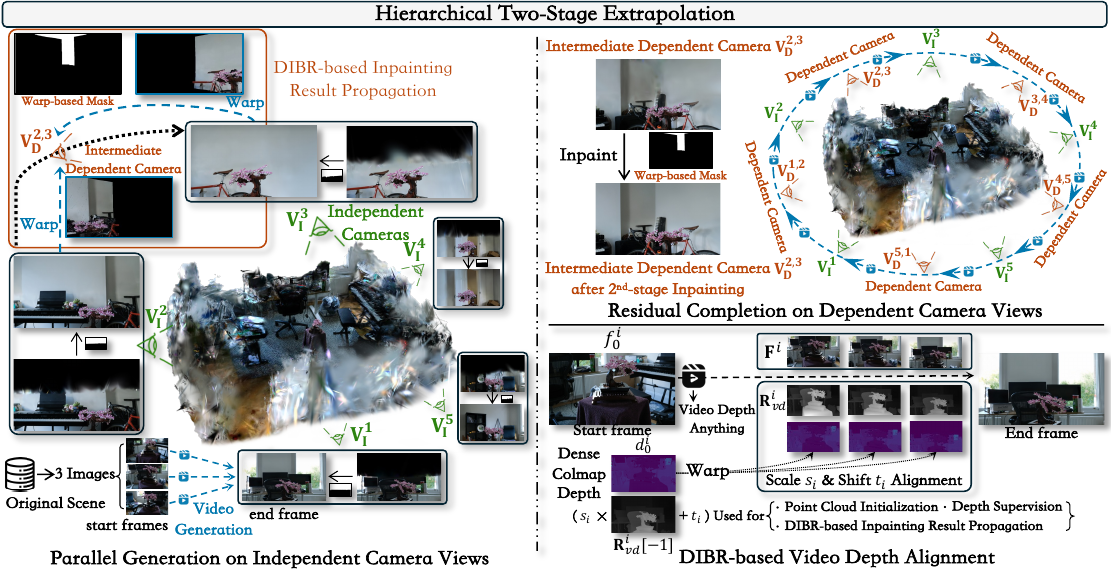}
    \caption{\textbf{An overview of our main pipeline.} It consists of extrapolations for \textcolor{idpdtgreen}{independent} and \textcolor{dpdtred}{dependent} camera views. The former achieves extrapolation on the original scene over large areas in a single pass. The latter performs a second-stage residual completion on the remaining un-extrapolated regions, enabling seamless scene extrapolation. DIBR-based video depth alignment provides robust depth priors throughout the process. Blue indicates the generated data used in each stage.}
    \label{fig:method_overview}
\end{figure*}

We present \textit{Filling the Unseen}, a hierarchical two-stage framework for large-scale and geometrically coherent scene extrapolation and interpolation. Figure~\ref{fig:method_overview} provides an overview of the full framework.

\textit{Filling the Unseen} is built upon a simple yet effective insight: a scene can be partitioned into mutually non-overlapping sub-regions to support parallel, conflict-free generation. These sub-regions are captured by a set of \textit{independent camera views} $\mathbf{V}_{\text{I}}$, identified via the mechanism described in Section~\ref{subsubsec:independent_camera_view_detection}. Leveraging this mechanism, we design the main pipeline in two stages. In the first stage (Section~\ref{subsubsec:extrapolation_for_independent_camera_views}), we perform parallel generation and reconstruction on $\mathbf{V}_{\text{I}}$, enabling large-scale scene expansion in a single pass and eliminating the need for repetitive generate–reconstruct–shift cycles. In the second stage (Section~\ref{subsubsec:extrapolation_for_dependent_camera_views}), we further process the remaining un-extrapolated gaps between views in $\mathbf{V}_{\text{I}}$, followed by a final interpolation step to achieve seamless scene extrapolation and interpolation. In Section~\ref{subsec:quality-aware_masks}, we introduce a novel quality-aware mask to take into consideration the inconsistency between generated content and the original scene, with the goal of promoting reliable updates while mitigating undesirable side-effects.

\subsection{Preliminary: Spherical Harmonics in 3DGS}
In 3DGS, Spherical Harmonics (SH) are employed as a set of orthonormal basis functions $Y$ to represent view-dependent color appearance.
Each 3D Gaussian stores a set of learnable SH coefficients $a_{l,m}$, where $l$ denotes the SH degree and $m$ indexes different directional modes. The $\mathbf{0}$th-order term $a_{0,0}$ captures the view-independent base color, and higher-order terms represent the view-dependent color. The final color $c(\mathbf{d})$ for the viewing direction $\mathbf{d}$ is computed by adding these two parts together:

{
\begin{equation}
c(\mathbf{d}) = 
\underbrace{\vphantom{\sum_{l=1}^{L}} a_{0,0} Y_0^0}_{c_{\text{base-color}}}
+ 
\underbrace{\sum_{l=1}^{L} \sum_{m=-l}^{l} a_{l,m} \, Y_l^m(\mathbf{d})}_{c_{\text{view-dependent-color}}(\mathbf{d})},
\qquad Y_0^0 = \frac{1}{2\sqrt{\pi}},
\end{equation}
}
where $Y_l^m$ denotes the SH basis function and $L$ denotes the maximum degree of SH.
This directional modeling capability is central to our QA-Mask design in Section~\ref{subsec:quality-aware_masks}.

\subsection{Independent Camera View Detection}
\label{subsubsec:independent_camera_view_detection}
Given a reconstructed 3D scene, the user first specifies a set of camera views $\mathbf{V}=\{\mathbf{V}^1, \mathbf{V}^2, ..., \mathbf{V}^N\}$ that indicates the regions of interest for extrapolation. We automatically identify a subset of independent camera views $\mathbf{V}_{\text{I}} \subseteq \mathbf{V}$, where two views $\mathbf{V}^i$ and $\mathbf{V}^j$ are considered \textit{independent} if their inpainted regions do not overlap. Independent camera views can be inpainted and reconstructed in parallel without conflicts. Detecting independence at the 2D image level is usually insufficient. Feature-based similarity metrics such as CLIP \cite{radford2021learning} or DINO \cite{oquab2023dinov2} are too coarse-grained for overlap detection, and feature matching methods \cite{lowe2004distinctive, sun2021loftr} fail on textureless surfaces like plain walls. We therefore operate in 3D, using mesh-based view frustum collision detection, as illustrated in Figure~\ref{fig:independent_camera_view_detection}.

The key idea is to construct a mesh-based view frustum that tightly encloses the 3D geometry of the inpainted region for each candidate view $\mathbf{V}^\text{i}$. If the frustums of $\mathbf{V}^\text{i}$ and $\mathbf{V}^\text{j}$ do not collide, their inpainted regions are guaranteed not to overlap, and the two views are independent. We proceed as follows:

\begin{enumerate}[leftmargin=17pt]
\item \textbf{Inpainting and depth estimation.} Inpaint each $\mathbf{V}^\text{i}$ with inpainting mask $\mathbf{M}^\text{i}$ to get $\mathbf{I}^\text{i, j}$. Then estimate each relative depth $\mathbf{R}^\text{i, j}$ using Depth Anything V2 \cite{yang2024depth}.

\item \textbf{Depth warping and alignment.} Find the closest camera pose $\mathbf{P}^\text{i}$ to $\mathbf{V}^\text{i}$ in the original training set, obtain the dense Colmap depth $\mathbf{D}^\text{i}$, and use depth image-based rendering (DIBR) to warp it to $\mathbf{V}^\text{i}$. Then align the warped depth with $\mathbf{R}^\text{i, j}$ to obtain a scale $s_{i, j}$ and an offset $t_{i, j}$:
{\small
\begin{equation}
    s_{i, j}, t_{i, j} = \arg\min\limits_{s,t}\sum {\begin{Vmatrix} Warp^{P^{i}\rightarrow V^{i}}(D^{i}) - R^{i, j} \end{Vmatrix}^2}.
\end{equation}}

\item \textbf{View frustum construction.} Use $s_{i, j}$ and $t_{i, j}$ to adjust $\mathbf{R}^\text{i, j}$ in the inpainted region:
\begin{equation}
    D^{i, j} = (s_{i, j} \times R^{i, j} + t_{i, j}) \times M^{i},
\end{equation}
and calculate the depth range $[\mathbf{D}^{\text{i, j}}_{\text{min}}, \mathbf{D}^{\text{i, j}}_{\text{max}}]$ for $D^{i, j}$. Next compute an average depth range $[\mathbf{D}^{\text{i}}_{\text{min}}, \mathbf{D}^{\text{i}}_{\text{max}}]$ across $\mathbf{n}$ inpainting results for $\mathbf{V}^\text{i}$:
\begin{equation}
    D^{i}_{min}=\frac{1}{n} \sum_{j=1}^{n} D^{i,j}_{min}, \: D^{i}_{max}=\frac{1}{n} \sum_{j=1}^{n} D^{i,j}_{max}.
\end{equation}
$[\mathbf{D}^{\text{i}}_{\text{min}}, \mathbf{D}^{\text{i}}_{\text{max}}]$ represents a high-probability depth range for the inpainting region of $\mathbf{V}^\text{i}$. We use $\mathbf{D}^{\text{i}}_{\text{min}}$ and $\mathbf{D}^{\text{i}}_{\text{max}}$ as the near and far planes to construct a view frustum for only the inpainting region of $\mathbf{V}^\text{i}$. This frustum encloses the 3D geometry of the inpainted results.

\item \textbf{Frustum collision detection.} After constructing frustums for each $\mathbf{V}^\text{i}$, we identify a set of independent camera views $\mathbf{V}_{\text{I}}$ via mesh-based collision detection. Next, we sort $\mathbf{V}_{\text{I}}$ based on the nearest neighbor and generate a closed camera view trajectory used for the following extrapolation.
\end{enumerate}

\subsection{Hierarchical Two-stage Scene Extrapolation and Interpolation}
\label{subsec:hierarchical_two-stage_scene_extrapolation}

\subsubsection{Parallel Generation on Independent Camera Views}
\label{subsubsec:extrapolation_for_independent_camera_views}

\leavevmode\par
\noindent In the first stage, we extrapolate the original scene on $\mathbf{V}_{\text{I}}$ over large areas in a single pass. We use a diffusion-based inpainting model \cite{zhang2023adding} for each $\mathbf{V}_{\text{I}}^{i}$ to fill up the black holes due to un-observations. Next we select 3 closest camera views to $\mathbf{V}_{\text{I}}^{i}$ from the original training set and generate 3 videos using \cite{xing2024dynamicrafter} in a way that one original training image being the start frame and the inpainted image of $\mathbf{V}_{\text{I}}^{i}$ being the end frame. These generated videos are used for the first stage extrapolation, as shown in the left part of Figure~\ref{fig:method_overview}.

However, seamless extrapolation cannot be achieved by the first stage alone. There is unfilled gap captured by the intermediate dependent camera view $\mathbf{V}_{\text{D}}^{i,i+1}$ between $\mathbf{V}_{\text{I}}^{i}$ and $\mathbf{V}_{\text{I}}^{i+1}$. We will perform extrapolation on $\mathbf{V}_{\text{D}}$ in the following second stage, which is the only one-time execution of the generate-reconstruct-shift cycle in our pipeline. Since image inpainting quality is highly related to the quality of the image itself, to ensure high inpainting quality for $\mathbf{V}_{\text{D}}$ in the next stage, the reconstruction in this stage needs to be robust to handle view shifting. Therefore, we aim to acquire as much multi-view information as possible during the first stage reconstruction. We employ DIBR-based warping to propagate the inpainted results of $\mathbf{V}_{\text{I}}$ to $\mathbf{V}_{\text{D}}$. Next, we introduce how to obtain accurate depth priors and perform propagations using the depths.

\noindent \textbf{DIBR-based Video Depth Alignment.} Estimating relative depth by \cite{yang2024depth} and scale-aligning it with SfM points are a common way \cite{chung2024depth} to acquire depth priors. However, it is inherently unstable and error-prone because of the sparsity and noise of SfM points. So we employ Video Depth Anything \cite{chen2025video} and DIBR to perform a robust depth alignment. As illustrated in the bottom right of Figure~\ref{fig:method_overview}, for each $\mathbf{V}_{\text{I}}^{i}$, we randomly choose one from the three generated videos and use \cite{chen2025video} to obtain relative video depth $\mathbf{R}_{\text{vd}}^i$. Since the start frame $f^i_{0}$ of the video is one original training image, we can warp its dense Colmap depth $d^i_{0}$ to all following frames $\mathbf{F}^i$, and then conduct a robust multi-image depth alignment using RANSAC and obtain the scale $s_{i}$ and offset $t_{i}$. We acquire depth $\mathbf{D}^{i}_{\text{I}}$ for the inpainted image of $\mathbf{V}_{\text{I}}^{i}$ by adjusting the last frame of $\mathbf{R}_{\text{vd}}^i$ with $s_{i}$ and $t_{i}$:
\begin{equation}
    s_i, t_i = \arg\min\limits_{s,t}\sum {\begin{Vmatrix} Warp^{f^i_{0}\rightarrow \mathbf{F}^i}(d^i_{0}) - \mathbf{R}_{\text{vd}}^i \end{Vmatrix}^2},
\end{equation}
\begin{equation}
    \mathbf{D}^{i}_{\text{I}} = s_{i} \times \mathbf{R}_{\text{vd}}^i[-1] + t_{i}.
\end{equation}
$\mathbf{D}^{i}_{\text{I}}$ can be used in extrapolation point cloud initialization, DIBR-based inpainting result propagation, and depth supervision.

\noindent \textbf{DIBR-based Inpainting Result Propagation.} We utilize depth priors $\mathbf{D}^{i}_{\text{I}}$ and $\mathbf{D}^{i+1}_{\text{I}}$ to warp the inpainted results on $\mathbf{V}_{\text{I}}^{i}$ and $\mathbf{V}_{\text{I}}^{i+1}$ to the intermediate dependent camera view $\mathbf{V}_{\text{D}}^{i,i+1}$ between $\mathbf{V}_{\text{I}}^{i}$ and $\mathbf{V}_{\text{I}}^{i+1}$, as shown in the upper left of Figure~\ref{fig:method_overview}. This provides additional supervision for improving the rendering quality of $\mathbf{V}_{\text{D}}^{i,i+1}$ to facilitate inpainting the remaining un-extrapolated gap in $\mathbf{V}_{\text{D}}^{i,i+1}$ at the second stage. Additionally, a warp-based mask indicating un-extrapolated gap is also acquired during DIBR-based warping to guide the later inpainting in the next stage.

\begin{figure}
    \centering
    \includegraphics[width=\linewidth]{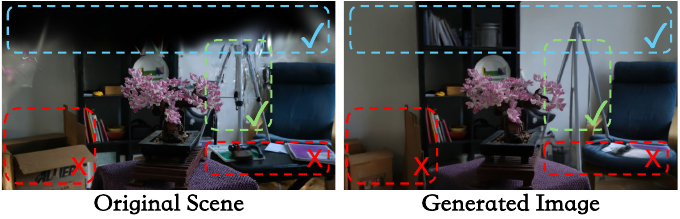}
    \caption{Under-reconstructed areas, including \textcolor{myblue}{holes} and \textcolor{mygreen}{low-quality regions}, should be updated using generated images. In contrast, well-reconstructed areas within the original scene should remain unchanged to avoid compromising the overall quality due to potential \textcolor{myred}{generative inconsistencies}.}
    \label{fig:why_oa_mask}
\end{figure}

\begin{figure}
    \centering
    \includegraphics[width=\linewidth]{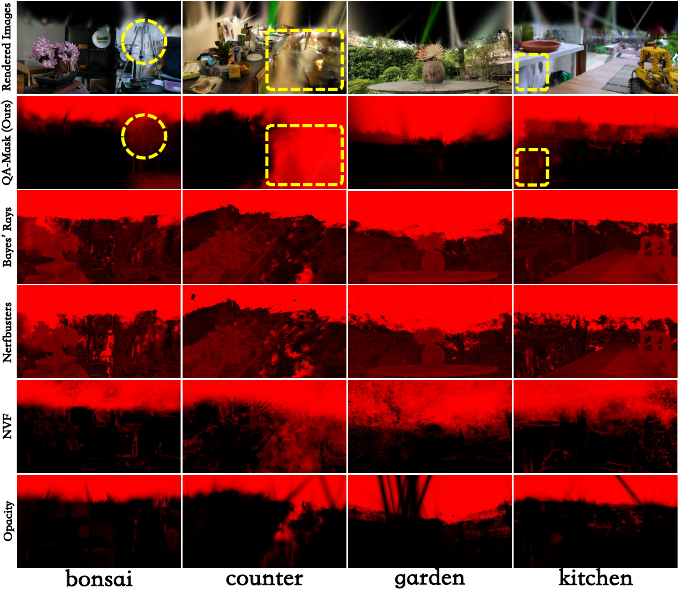}
    \caption{Visualizations of rendered RGB images and QA-Mask from extrapolation views on four scenes in Mip-NeRF 360. QA-Mask identified holes and low rendering quality regions (dashed yellow) that align well with human perception. However, Bayes' Rays and Nerfbusters fail to clearly distinguish between low-quality and well-reconstructed areas, while opacity maps and NVF are unable to accurately identify the low-quality areas within the original scene or maintain high (uncertainty) values across the hole areas.}
    \label{fig:ood_mask_vis}
\end{figure}

\begin{figure*}[t]
    \centering
    \includegraphics[width=\linewidth]{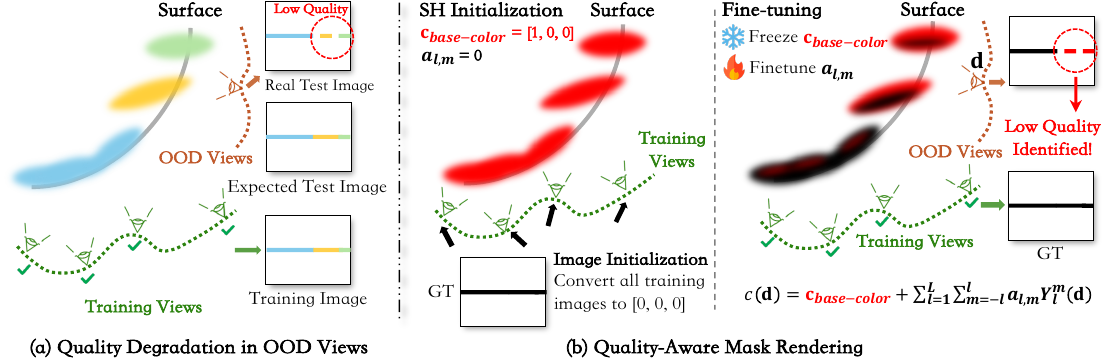}
    \caption{(a) Rendering degradations on out-of-distribution views. 3D Gaussians overfit on training views but underfit on out-of-distribution (OOD) views which results in degraded rendering. (b) Quality-Aware Mask rendering method. We first prepare a pretrained 3DGS model by setting the SH base color to red, view-dependent color coefficients $a_{l,m}$ to 0 and converting all original training images to black. We fine-tune this model with only $a_{l,m}$ receiving gradients. After fine-tuning, within the training view distribution with high rendering quality, the base color is canceled out by $a_{l,m}$ with the final color being black. While in OOD views with low rendering quality, the lack of black supervisions causes Gaussians to remain the base color.}
    \label{fig:ood_mask}
\end{figure*}

\subsubsection{Residual Completion on Dependent Camera Views}
\label{subsubsec:extrapolation_for_dependent_camera_views}
\leavevmode\par
\noindent In the second stage, we first inpaint each intermediate dependent camera view $\mathbf{V}_{\text{D}}^{i,i+1}$. Then every independent or dependent camera view in the closed extrapolation trajectory corresponds to a complete image without un-extrapolated regions. Next, as illustrated in the upper right of Figure~\ref{fig:method_overview}, we generate a video for every two neighboring views in the trajectory with one being the start frame and the other being the end frame. These videos are used for the second stage extrapolation.

With these two stages, the original scene has been extrapolated at a large scale along the extrapolation trajectory.

\subsubsection{Scene Interpolation via Video Reuse}
\label{subsubsec:scene_interpolation}
\leavevmode\par
\noindent Scene interpolation can improve overall rendering quality in the original observed regions, which suffer from degradations on views from the extrapolation trajectory, as shown in Figure~\ref{fig:teaser}. We exploit two sources of interpolation data. First, the videos generated during extrapolation inherently contain intermediate frames between original training images and inpainted views, providing multi-view supervision for the original scene. Second, after stage 2 extrapolation, we render images along the closed extrapolation trajectory and pass them through the artifact removal model \cite{wu2025difix3d+}, whose outputs are added to the training set on-the-fly. Both strategies are applied jointly during this final interpolation stage.

\subsection{Quality-Aware Mask Rendering}
\label{subsec:quality-aware_masks}
When updating a scene with generated content, inconsistency between the generated data and the original scene often occurs. This necessitates selective utilization on the generated data. A reasonable strategy is to spatially re-weight loss functions based on current rendering quality. As exemplified in Figure~\ref{fig:why_oa_mask}, in areas without Gaussians or with low rendering quality, generated data should be applied for extrapolation and interpolation. While in well-reconstructed areas, they should not be used, to prevent data inconsistency from corrupting the original scene. Based on this idea, in this section, we propose our novel method to render Quality-Aware Mask (QA-Mask).

The core of 3DGS is to find one feasible solution for a set of 3D Gaussian primitives under multi-view constraints. This solution overfits on training views but underfits on out-of-distribution novel views. In other words, rendering quality is highly related to the viewing directions (Figure~\ref{fig:ood_mask} (a)). Since Spherical Harmonics (SH) possesses directional modeling ability, we exploit SH to model rendering quality. 

Next, we first outline the steps to obtain QA-Mask from a 3DGS model pretrained on the original scene, as illustrated in Figure~\ref{fig:ood_mask} (b), and then explain how it works:

\begin{enumerate}[leftmargin=17pt]
\item \textbf{SH coefficient initialization.} We set $c_\text{base-color}$ to one arbitrary color from RGB, e.g., red in our example, and the view-dependent coefficients $a_{l,m}$ to 0.
\item \textbf{Training image initialization.} We convert original training images to all black with RGB of $(0, 0, 0)$.
\item \textbf{Fine-tuning.} We fine-tune this 3DGS model on converted black training images with only $a_{l,m}$ receiving gradients.
\end{enumerate}

During fine-tuning, the Gaussians with the fixed red base color rely on view-dependent $a_{l,m}$ to shift the final color from red to black. However, on out-of-distribution novel views that are far from training supervisions, since without black signals, the final SH color remains largely red. At its core, we use SH to model “which directions are supervised.” SH colors such as black and red in this example, can then serve as quantitative indicators for how well a pixel is supervised from this direction, i.e., the rendering quality. We show some visualization examples in Figure~\ref{fig:ood_mask_vis}, where the background color is set to red during rendering. It is worth noting the choice of red/black is arbitrary and any two distinguishable colors suffice. We verified it with alternative colors in the supplementary.

After normalizing the R channel of QA-Mask $\mathbf{M}$, they are employed as per-pixel learning weights applied to our reconstruction loss, emphasizing under-reconstructed areas while suppressing updates in well-reconstructed areas:

\begin{equation}
\small
\mathcal{L}
=
\lambda_{1}\lVert\mathbf{M}\odot(\mathbf{I}_{\mathrm{p}}-\mathbf{I}_{\mathrm{gt}})\rVert_{1}
+\lambda_{2}\mathrm{SSIM}(\mathbf{M}\odot\mathbf{I}_{\mathrm{p}},\mathbf{M}\odot\mathbf{I}_{\mathrm{gt}})
+\lambda_{3}\lVert\mathbf{M}\odot(\mathbf{D}_{\mathrm{p}}-\mathbf{D}_{\mathrm{gt}})\rVert_{1}.
\end{equation}
where $\mathbf{I}_{\text{p}}$, $\mathbf{I}_{\text{gt}}$ denotes rendered and ground truth images, $\mathbf{D}_{\text{p}}$, $\mathbf{D}_{\text{gt}}$ denotes rendered and ground truth depths.

\section{Experiments}
\label{sec:experiments}

\begin{figure*}
    \centering
    \includegraphics[width=\linewidth]{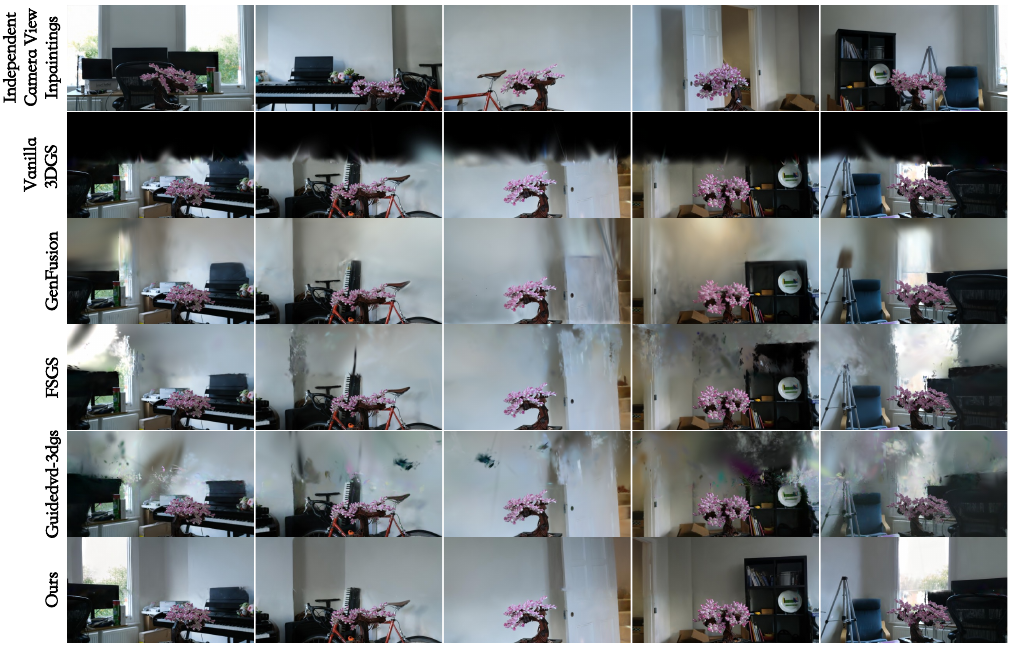}
    \caption{Qualitative comparisons of extrapolation on Mip-NeRF 360. The first row shows the inpainted images on independent camera views $\mathbf{V}_{\text{I}}$. The following rows show the images rendered on the camera views between each pair of $\mathbf{V}_{\text{I}}^{i}, \mathbf{V}_{\text{I}}^{i+1}$. Our framework achieves high rendering quality on extrapolated areas while also leaves well-reconstructed original areas untouched.}
    \label{fig:mipner360_fig}
\end{figure*}

\subsection{Experimental Settings}
\label{baselines_exp}

\noindent \textbf{Datasets.}
We conduct experiments on two datasets Mip-NeRF 360 \cite{barron2022mip} and ScanNet++ \cite{yeshwanthliu2023scannetpp}: 4 scenes from Mip-NeRF 360 (bonsai, counter, garden, kitchen) and 10 scenes from ScanNet++.

\noindent \textbf{Evaluation Settings.}
Since there are no ground-truth images for extrapolation viewpoints in Mip-NeRF 360 by construction, reference metrics (PSNR/SSIM/LPIPS) are inapplicable. We therefore adopt non-reference metrics including \textbf{CLIP-IQA} \cite{wang2022exploring}, \textbf{QualiCLIP} \cite{agnolucci2024qualityaware}, matching-based Multi-View Consistency (\textbf{MVC}) \cite{detone2018superpoint, sarlin2020superglue}, and CLIP-based Semantic Similarity (\textbf{SemSim})  \cite{radford2021learning} as proxies for perceptual quality, multi-view consistency, and semantic fidelity. 

For ScanNet++, we design a controlled evaluation protocol to enable quantitative assessment using reference metrics, including \textbf{PSNR}, \textbf{SSIM}, and \textbf{LPIPS}.
To implement this, we reconstruct only half of each scene, and treat the other half as the unseen extrapolation target, where original training
images serve as ground truth for reference-metric evaluation. 
Details about metrics and settings can be found in the supplementary.

\begin{table}[!tbp]
\centering
\caption{Non-reference metrics and runtime on Mip-NeRF 360. C-IQA denotes CLIP-IQA. QUALI-C denotes QUALI-CLIP.}
{
\setlength{\tabcolsep}{1.5pt}
\begin{tabular}{c|ccccc}
\toprule
\textbf{Methods} & \textbf{C-IQA}$\uparrow$ & \textbf{QUALI-C}$\uparrow$ & \textbf{MVC}$\uparrow$ & \textbf{SemSim}$\uparrow$ & \textbf{Time}\\
\midrule
Few-Shot GS      & 0.278 & 0.352 & 1792 & 0.707 & 24min\\
DiFix3D+          & 0.354 & 0.451 & 1755 & 0.687 & 53min\\
GenFusion        & 0.340 & 0.378 & 1157 & 0.712 & 44min\\
Guidedvd-3dgs    & 0.230 & 0.298 & 1845 & 0.691 & 306min\\

\midrule
\textbf{Ours}    & {\textbf{0.404}} & {\textbf{0.498}} & {\textbf{1972}} & {\textbf{0.737}} & 77min\\
\bottomrule
\end{tabular}
}
\label{tab:mipnerf360_tab}
\end{table}
\begin{table}[!tbp]
\centering
\caption{Comparisons of reference metrics on ScanNet++. Our method achieves the best performance averaged on 10 scenes.}
\resizebox{0.75\linewidth}{!}{
\begin{tabular}{c|ccc}
\toprule
\textbf{Methods} & \textbf{PSNR}$\uparrow$ & \textbf{SSIM}$\uparrow$ & \textbf{LPIPS}$\downarrow$ \\
\midrule

Few-Shot GS & 18.92 & 0.80 & 0.3438 \\

DiFix3D+ & 20.93 & 0.84 & 0.2554 \\

GenFusion & 21.93 & 0.85 &  0.2502 \\

Guidedvd-3dgs & 22.10 & 0.84 & 0.2662 \\

\midrule
\textbf{Ours} & \textbf{23.79} & \textbf{0.88} & \textbf{0.2300} \\
\bottomrule

\end{tabular}
}
\label{tab:scannetpp_main_tab}
\end{table}

\subsection{Baselines and Implementation Details}
\noindent \textbf{Baselines.} 
We compare with recent SOTA methods: Few-Shot Gaussian Splatting (FSGS) \cite{zhu2024fsgs}, DiFix3D+ \cite{wu2025difix3d+}, GenFusion \cite{wu2025genfusion}, and Guidedvd-3dgs \cite{zhong2025taming}. 
For experiments on Mip-NeRF 360, we supply FSGS and DiFix3D+ with the same inpainted images on independent camera views produced by our framework. As these methods are not explicitly designed for scene extrapolation, we adapt their settings accordingly to enable a meaningful comparison.

\noindent \textbf{Implementation Details.}
We use $\lambda_{1}=0.7$, $\lambda_{2}=0.2$, $\lambda_{3}=0.1$ in our experiments.
When performing reconstructions, our method is trained for 30,000 iterations for each stage. For QA-Mask, we use 5-degree SH and fine-tune for 20,000 iterations. All experiments are conducted on a NVIDIA RTX A6000 GPU.

\subsection{Comparisons on Non-reference Setting}
The non-reference metrics and runtime evaluated on Mip-NeRF 360 are reported in Table~\ref{tab:mipnerf360_tab}. Our method outperforms these baselines over all metrics at an acceptable time cost, demonstrating the superiority of our method in terms of perceptual image quality, multi-view consistency, and high-level semantic consistency. We show qualitative comparisons in Figure~\ref{fig:mipner360_fig}. Our method enjoys high rendering quality and multi-view consistency in extrapolated regions compared to baselines that are plagued by blurring and floaters. This stems from our hierarchical two-stage extrapolation design that effectively prevents the compounding of errors observed in prior approaches. In the same time, no sign of rendering deterioration can be observed within original areas for our method thanks to the proposed QA-Mask. While other models such as GenFusion and Guidedvd-3dgs suffer from quality drops in these areas.

The runtime for our method in Table~\ref{tab:mipnerf360_tab} includes the two-stage extrapolation pipeline, interpolation, and the training of QA-Mask. More images and videos are provided in the supplementary.

\subsection{Comparisons on Reference Setting}
\label{sec:comparisons_on_reference_setting}
Quantitative results for reference metrics on ScanNet++ are reported in Table~\ref{tab:scannetpp_main_tab}. Qualitative comparisons are shown in the supplementary, where our method delivers noticeably better visual quality in both extrapolated and existing regions. Other methods exhibit artifacts and geometric distortions to varying extents, despite the availability of partial ground-truth images in this setting. More images and videos on ScanNet++ are provided in the supplementary.

\subsection{Generalizability of QA-Mask}
\label{Generalizability_vufield}
\begin{figure}[htbp]
    \centering
    \includegraphics[width=\linewidth]{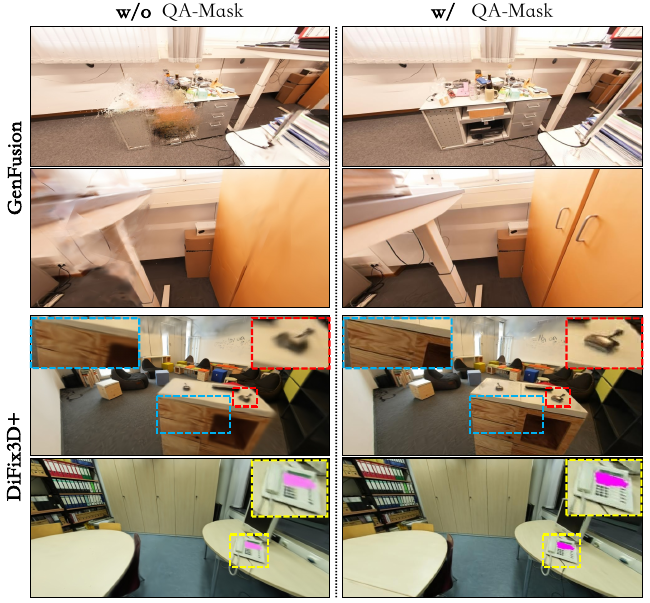}
        \caption{Comparison between w/ and w/o QA-Mask on GenFusion and DiFix3D+. QA-Mask prevents generated data from disrupting well-reconstructed areas in the original scene. Noticeable improvements can be observed for both models.}
    \label{fig:ood_mask_generalization_fig}
\end{figure}

In this section, we compare our QA-Mask with a couple of baselines including opacity maps, Bayes' Rays \cite{goli2024bayes}, Nerfbusters \cite{warburg2023nerfbusters}, and Neural Visibility Field (NVF) \cite{xue2024neural} by plugging them into DiFix3D+ and GenFusion, and carry out a comparative study on ScanNet++. We report PSNR and LPIPS for the extrapolated and existing regions separately in Table~\ref{tab:ood_mask_generalization_tab}. After introducing QA-Mask, the metrics of the existing regions have been improved for both DiFix3D+ and GenFusion, which demonstrates that QA-Mask can prevent generated data from compromising the original scene. Meanwhile, extrapolated regions maintain comparable rendering quality than the counterpart without QA-Mask, proving that QA-Mask does not hinder desired scene updates using generated data. This appealing feature of our QA-Mask, which we did not observe in other baselines as shown in Figure~\ref{fig:ood_mask_vis}, makes it serve as a plug-and-play enhancement module for a series of generation-based reconstruction models. Qualitative comparative examples are shown in Figure~\ref{fig:ood_mask_generalization_fig}.

\begin{table}[!tbp]
\centering
{
\caption{Generalizability of QA-Mask on ScanNet++. Exst. denotes existing and Extrapltd. denotes extrapolated.}

\label{tab:ood_mask_generalization_tab}
\begin{tabular}{c|cc|cc}
\toprule
\multirow{2}{*}{\textbf{Methods}} 
& \multicolumn{2}{c|}{\textbf{PSNR}$\uparrow$} 
& \multicolumn{2}{c}{\textbf{LPIPS}$\downarrow$} \\
& Exst. & Extrapltd. & Exst. & Extrapltd. \\ 
\midrule
{\cellcolor{gray!20}DiFix3D+}
& {\cellcolor{gray!20}28.11} & {\cellcolor{gray!20}\textbf{10.23}} & {\cellcolor{gray!20}0.1187} & {\cellcolor{gray!20}0.4654} \\

\midrule
w/ Bayes' Rays
& 27.87
& 10.20
& 0.1235
& 0.4701 \\

\midrule
w/ Nerfbusters
& 27.73
& 10.09
& 0.1193
& 0.4659 \\

\midrule
w/ NVF
& 28.19
& 10.02
& 0.1093
& 0.4700 \\

\midrule
w/ Opacity Maps
& 28.15
& 10.17
& 0.1109
& 0.4655 \\

\midrule
w/ QA-Mask (Ours)
& \textbf{28.58}
& 10.20
& \textbf{0.1075}
& \textbf{0.4653} \\

\midrule\midrule
{\cellcolor{gray!20}GenFusion}
& {\cellcolor{gray!20}26.49} & {\cellcolor{gray!20}15.04} & {\cellcolor{gray!20}0.1367} & {\cellcolor{gray!20}\textbf{0.4156}} \\

\midrule
w/ Bayes' Rays
& 27.79
& 14.99
& 0.1146
& 0.4161 \\

\midrule
w/ Nerfbusters
& 28.01
& 15.00
& 0.1203
& 0.4176 \\

\midrule
w/ NVF
& 28.07
& 14.77
& 0.1121
& 0.4203 \\

\midrule
w/ Opacity Maps
& 28.03
& 15.04
& 0.1296
& 0.4186 \\

\midrule
w/ QA-Mask (Ours)
& \textbf{28.33}
& \textbf{15.10}
& \textbf{0.1056}
& 0.4193 \\
\bottomrule
\end{tabular}
}
\end{table}

\subsection{Ablation Study}
\begin{table}[!tbp]
\centering
\caption{Ablation study for core components and hyperparameter selection on both Mip-NeRF 360 and ScanNet++. Idpdt Cam Detection denotes Independent Camera View Detection.}
{
\begin{tabular}{c|ccc}
\toprule
{\cellcolor{gray!20}Mip-NeRF 360} & {\cellcolor{gray!20}\textbf{CLIP-IQA}$\uparrow$} & {\cellcolor{gray!20}\textbf{MVC}$\uparrow$} & {\cellcolor{gray!20}\textbf{SemSim}$\uparrow$} \\
\midrule
w/o QA-Mask       & 0.343 & 1673 & 0.719 \\
w/o DIBR Propagation          & 0.316 & 1468 & 0.697 \\
Ours  & \textbf{0.404} & \textbf{1972} & \textbf{0.737} \\

\midrule
{\cellcolor{gray!20}ScanNet++} 
& {\cellcolor{gray!20}\textbf{PSNR}$\uparrow$} 
& {\cellcolor{gray!20}\textbf{SSIM}$\uparrow$} 
& {\cellcolor{gray!20}\textbf{LPIPS}$\downarrow$} \\

w/o Idpdt Cam Detection & 20.27 & 0.80 & 0.3161 \\
w/o Depth Supervision & 23.66 & \textbf{0.88} & 0.2302 \\
w/ Degree-3 SH QA-Mask & 22.45 & 0.84 & 0.2432 \\
w/ Degree-4 SH QA-Mask & 23.28 & 0.86 & 0.2331 \\
Ours & \textbf{23.79} & \textbf{0.88} & \textbf{0.2300} \\

\bottomrule
\end{tabular}
}
\label{tab:supp_ablation_tab}
\end{table}
In this section, we perform ablation studies for core components and hyperparameter selection on both Mip-NeRF 360 and ScanNet++, and present the results in Table~\ref{tab:supp_ablation_tab}. Experiments on Mip-NeRF 360 showed that removing QA-Mask or DIBR-based propagation both yields worse performance. Since QA-Mask is designed to suppress unintended updates without impeding desired extrapolation and interpolation. DIBR-based propagation leverages depth priors to provide additional multi-view cues for the first stage reconstruction so that facilitates the transition to the next stage.

We also conducted experiments on ScanNet++. First by removing the independent camera view detection mechanism and evenly choosing the same number of camera views in a conflict-agnostic manner, the rendered results are severely plagued by content conflicts reflecting as considerable performance drop. Second, the w/o depth supervision variant achieved slightly lower performance, indicating our video depth alignment module provides accurate depth information. Additionally, we quantitatively justified our choice for the 5-degree SH in QA-Mask over 3 or 4-degree, where high degree enhances SH's ability to capture high-frequency angular variations, enabling more precise directional modeling.

\section{Conclusion}
We introduce \textbf{\textit{Filling the Unseen}}, a holistic framework for large-scale scene extrapolation and interpolation based on 3D Gaussian Splatting. By first detecting independent camera views that support parallel generation and reconstruction, we redesign a hierarchical two-stage extrapolation pipeline and achieved high-quality scene extrapolation. We hope this hierarchical pipeline can provide new design perspective on scene generation tasks for the community. Furthermore, the proposed QA-Mask effectively prevents generated content from negatively impacting the well-reconstructed areas of original scenes, which can serve as a plug-and-play enhancement module for a series of generation-based reconstruction models.

%%
%% The next two lines define the bibliography style to be used, and
%% the bibliography file.
\bibliographystyle{ACM-Reference-Format}
\balance
\bibliography{sample-base}

\clearpage
\appendix

\end{document}